\documentclass{article}

\usepackage[preprint]{neurips_2024}

\usepackage[utf8]{inputenc} 
\usepackage[T1]{fontenc}    
\usepackage{hyperref}       
\usepackage{url}            
\usepackage{booktabs}       
\usepackage{amsfonts}       
\usepackage{nicefrac}       
\usepackage{microtype}      
\usepackage{xcolor}         

\usepackage{amsmath}
\usepackage{amssymb}
\usepackage{amsthm}
\usepackage{subcaption}
\usepackage{multirow}
\usepackage{booktabs}

\usepackage{url}            
\usepackage{booktabs}       
\usepackage{amsfonts}       
\usepackage{nicefrac}       
\usepackage{xcolor}         
\usepackage{enumitem}
\usepackage{graphicx}       
\usepackage{wrapfig}        
    \usepackage[colorinlistoftodos]{todonotes}

\usepackage{algorithm}
\usepackage{algorithmic}

\theoremstyle{plain}
\newtheorem{theorem}{Theorem}[section]

\theoremstyle{definition}
\newtheorem{definition}[theorem]{Definition}

\theoremstyle{remark}
\newtheorem{remarks}[theorem]{Remarks}

\theoremstyle{plain}

\newcommand{\R}{\mathbb{R}}

\newcommand{\I}{\mathcal{I}}

\newcommand{\comment}[1]{}

\title{STAT: Shrinking Transformers After Training}

\author{%
  Megan Flynn
    \\
  Department of Physics\\
  Cornell University\\
  Ithaca, NY 14853 \\
  \texttt{mr2268@cornell.edu} \\
  \And
  Alexander Wang
    \\
  Department of Computer Science\\
  Cornell University\\
  Ithaca, NY 14853 \\
  \texttt{aw576@cornell.edu} \\
  \And
  Dean Edward Alvarez
    \\
  Department of Computer Science\\
  University of Illinois Urbana-Champaign\\
  Urbana, IL 61801 \\
  \texttt{deana3@illinois.edu} \\
  \And
  Christopher De Sa
    \\
  Department of Computer Science\\
  Cornell University\\
  Ithaca, NY 14853 \\
  \texttt{cmd353@cornell.edu} \\
  \And
  Anil Damle
    \\
  Department of Computer Science\\
  Cornell University\\
  Ithaca, NY 14853 \\
  \texttt{damle@cornell.edu} \\
}

\begin{document}

\maketitle

\begin{abstract}
We present STAT: a simple algorithm to prune transformer models without any fine-tuning.
STAT 
eliminates both attention heads and neurons from the network, while preserving accuracy by calculating a correction to the weights of the next layer. 
Each layer block in the network is compressed using a series of principled matrix factorizations that preserve the network structure. 
Our entire algorithm takes minutes to compress BERT, and less than three hours to compress models with 7B parameters using a single GPU.  
Using only several hundred data examples, STAT preserves the output of the network and improves upon existing gradient-free pruning methods. It is even competitive with methods that include significant fine-tuning. 
We demonstrate our method on both encoder and decoder architectures, including BERT, DistilBERT, and Llama-2 using benchmarks such as GLUE, Squad, WikiText2. 
\end{abstract}

\section{Introduction}
Transformer models have become ubiquitous across many fields, especially natural language processing. As such, the amount of resources devoted to training them from start to finish has increased rapidly. It has become common for highly resourced groups to train general-purpose models on massive datasets and then to release these models to the community open-source. This practice allows others to specialize these models for their particular use cases \cite{Devlin_BERT, Sanh_DistilBERT, zhang2022opt}. However, the best-performing models are incredibly large, and often require compression (e.g., pruning and/or quantization) before they can be used by many practitioners.

It is often easiest to realize real-world speedups (with the least engineering effort) when using structured pruning methods, especially when the ``structure'' means removing entire neurons, attention heads, layers, or compressing the embedding layer---these actions leave the network structure unchanged and just make its constituent pieces smaller.  While quantization methods can typically greatly compress networks, realizing the theoretical gains in practice can be difficult when, e.g., methods compress to non-standard floating point lengths that are not supported in hardware. Similarly, unstructured pruning methods can also achieve large compression ratios, but require either specialized hardware or data structures to realize performance gains.  

Several structured pruning methods have been studied for transformers, some which prune during training and some which train and then prune.  However, many require either significant re-training after pruning, or a very large amount of resources to prune and compute corrections.  

We propose STAT, a new structured pruning method that eliminates heads and neurons from the attention blocks and fully connected blocks of the network---all without requiring any fine-tuning. To accomplish this task we build on the methodology from~\citet{chee2022model} and use pivoted QR factorizations applied to the activation output for a small amount of (unlabeled) data at intermediary parts of the network to select heads and neurons to eliminate. 
We calculate corrections using the same data, on a single GPU.  
This enables us to minimize the error produced by pruning the network without resorting to expensive fine tuning. We can leverage randomized methods in numerical linear algebra (with a slight twist given our specific setting to enable additional parallelism) to scale the method up to modern models.

We give a basic overview of interpolative decompositions, the backbone of our method, and their computation in Section \ref{ID}.  In Section \ref{Application}, we show how to leverage interpolative decompositions to prune fully connected layers and attention heads.  We provide experimental results for BERT, DistilBERT and Llama in Section \ref{Experiment}, and perform a brief ablation analysis of our method in Section \ref{Ablation}.

\subsection{Summary of Contributions}
\begin{enumerate}
    \item We provide a method to compress transformer networks that slims the network by pruning attention heads and neurons while updating the subsequent layer to compensate---allowing us to  preserve accuracy without fine tuning. 
    \item We illustrate the effectiveness of our method using encoder-based networks such as BERT \cite{Devlin_BERT} and DistilBERT \cite{Sanh_DistilBERT}, and significantly outperform SOTA methods by providing a better FLOPS/Accuracy tradeoff without fine tuning. In fact, we often outperform methods that rely on significant fine tuning
    \item We demonstrate that using randomized techniques enables our methodology to scale by compressing the much larger modern generative decoder model Llama-2 7B \cite{touvron2023llama} in a few hours on a single GPU.  
\end{enumerate}

\subsection{Related Work}

Reductions in the memory footprint and inference speed of pre-trained transformer models can be accomplished through knowledge distillation
\cite{jiao2020tinybert, sanh2020distilbert, wilson2021kd, sun2019patient, wang2020minilm}, quantization \cite{kim2021ibert, Shen_Dong_Ye_Ma_Yao_Gholami_Mahoney_Keutzer_2020, Zadeh_2020, Zafrir_2019}, pruning \cite{kurtic2022optimal, Sajjad_2023, hou2020dynabert, lagunas2021block, frantar2023sparsegpt, parnami2021pruning} or other methods such as matrix factorization \cite{Wang_2020, lin2020pruning}. We focus on methods which require no (or minimal) re-training. 

\paragraph{Transformer Pruning}
Pruning has been a fruitful method to eliminate insignificant weights in neural networks. It can be split into two main categories: structured and unstructured pruning. Previous works have applied several unstructured pruning methods to transformers including magnitude \cite{gale2019state}, first order \cite{sanh2020movement}, and second order \cite{kurtic2022optimal} pruning. However, it is difficult to leverage unstructured sparsity to speed up models with commercially available hardware. 

As a result, several structured pruning methods have been proposed to eliminate groups of parameters at a time. 
Low-rank factorization \cite{Wang_2020}, block-wise sparsity \cite{li2020efficient}, and tile-wise sparsity \cite{guo2020accelerating} were studied to prune structured sets of parameters in weight matrices. 
Researchers have investigated how to specialize pruning methods for transformers, including
how coarse-grained structures like attention heads \cite{michel2019sixteen, voita2019analyzing, parnami2021pruning}, channels \cite{he2017feat}, and entire layers \cite{shim2021layerwise, fan2019reducing, Sajjad_2023} could be pruned away.


While structured methods may achieve significant speedups, they often require re-training or knowledge distillation to recover accuracy.\cite{lagunas2021block, xia2022structured}
In some cases this can exceed the resource consumption of training the original network from scratch, especially when the pruning pipeline also introduces new hyperparameters. \cite{hou2020dynabert, lagunas2021block, Liu_Lin_Yuan_2021}

\paragraph{Retraining Free Transformer Compression}


Both structured \cite{kim2020neuron, srinivas2015datafree, yvinec2021red} and unstructured \cite{lazarevich2021posttraining, frantar2022spdy, mussay2020coreset} post-training pruning schemes have been examined for CNNs and fully connected layers. 
However, CNNs have repeating linear layers and element-wise nonlinearity, whereas transformers have multi-head attention layers, which makes many of those methods nontransferable. 

One method presented by \citet{rtfp} leverages a block-diagonal approximation of the Fischer information matrix to inform which attention heads to prune. Their method achieves a 40\% reduction in FLOPs while maintaining within 1\% accuracy of the baseline model for several GLUE and SQuAD tasks. However, the approximation used by this work is limited in its efficacy and, as we will show, our principled use of structured matrix factorizations yields a favorable trade off between accuracy and model size.
Another method presented in Tukan et al. \cite{tukan2021nofinetuning} compresses the embedding layer of the network using low-rank approximation.  

Three major methods have come out in the past year which present re-training free results on truly large language models.  The first, LLM Surgeon, \cite{ouderaa2024the}, uses both activations and gradients to estimate the curvature of the loss function and therefore select structures to remove and update remaining weights to compensate. Their multi-shot algorithm provides very accurate state of the art results, but requires a large amount of computational resources to reach its full potential.  
The second, SliceGPT, \cite{ashkboos2024slicegpt}, uses principal component analysis to compress the network along the embedding dimension, based on the observation that layer norms can be re-written as RMS norms, which commutes with multiplication by unitary matrices.  
The third, WANDA, compresses the network by deleting weights using the activations as a heuristic.  However, it does not use a correction matrix and only produces results for unstructured and 2:4 structured sparsity.

\paragraph{Quantization}
Post-training Quantization as a means to reduce the memory footprint of neural networks while retaining accuracy has shown promise \cite{kim2021ibert, Zadeh_2020, Zafrir_2019, Shen_Dong_Ye_Ma_Yao_Gholami_Mahoney_Keutzer_2020}.
The simplest method, round-to-nearest, often produces reasonable results at moderate levels of compression \cite{nagel2020down, lagunas2021block}.  
Other methods \cite{pmlr-v119-wang20c, hubara2020improving, chee2023quip,egiazarian2024extreme,shao2024omniquant} use data to calibrate quantization, and have begun pushing towards 3 bit and even 2 bit quantization for large networks.
 Nevertheless, there are challenges in realizing inference speedups using arbitrary quantization and we consider this work complementary to our own (with different strengths and weaknesses) rather than a direct alternative.
\section{Interpolative decompositions}
Similarly to the methodology presented in~\citet{chee2022model}, a core component of our compression technique is the structured matrix factorization known as an interpolative decomposition (ID). In particular, IDs provide a principled way to subselect neurons (in fully connected layers) and heads (in attention layers); moreover, they provide a ``correction'' we can fuse into the next layer to retain accuracy. The key feature of an ID for this purpose is that it builds a low-rank approximation of a matrix using columns of the matrix itself---these selected columns will inform which neurons/heads to keep. Notably, while the singular value decomposition (SVD) provides an optimal low-rank approximation, it is inadequate for our task because singular vectors need not be columns of the matrix itself and, therefore, do not directly provide information for structured pruning.

\paragraph{Definitions and computation}
IDs have received considerable attention~\cite{cheng2005compression,martinsson2011randomized,goreinov1997theory}, particularly within the field of rank-structured matrices~\cite{ho2012fast,ho2015hierarchical,ho2016hierarchical,martinsson2005fast,martinsson2019fast,minden2017recursive}. Moreover, they are closely related to CUR decompositions~\cite{mahoney2009cur,voronin2017efficient} and subset selection problems~\cite{boutsidis2009improved,civril2009selecting,tropp2009column}. Because they are key to our methodology we provide a brief overview here along with notes about how they are computed.

\begin{definition}[Interpolative Decomposition]
\label{def:ID}
Consider $A \in \R^{n \times m},$ for any $\epsilon \geq 0$ an $\epsilon$-accurate \emph{interpolative decomposition} is a factorization $A \approx A_{:,\I} T,$ where $A_{:,\I}$ is a subset of columns of $A$ associated with $\I \subset [m]$ and $T$ is an interpolation matrix such that $\|A - A_{:,\I} T \|_2 \leq \epsilon\|A\|_2.$
\end{definition}

\begin{remarks}
\hspace{-0.8em}
In practice, we want to minimize $k\equiv\lvert\I\rvert$ while ensuring the accuracy requirement, and we would like $T$ to have entries of reasonable magnitude for stability. Ideally the error will be close to the best possible for a given $k$ (i.e., $\|A - A_{:,\I}T\|_2 = c \sigma_{k+1}(A)$ for some small $c \geq 1$).
\end{remarks}

Computing an ``optimal'' ID (i.e., minimizing the error for a given $k$) is closely related to the provably hard problem of computing maximum volume subsets~\cite{civril2009selecting}. Nevertheless, one of the reasons IDs have found extensive use is that (strong) rank-revealing QR factorizations~\cite{businger1965linear,chan1992some,chandrasekaran1994rank,gu1996efficient,hong1992rank} provide practical ways to compute useful IDs. A rank-revealing QR factorization of a matrix $A \in \R^{n \times m}$ for some $k < \min(n,m)$ computes a permutation $\Pi$, a matrix with orthonormal columns $Q$, and an upper-trapezoidal matrix $R$, such that 
\begin{equation}
\label{eq:rankQR}
    A
    \begin{bmatrix}
    \Pi_1 & \Pi_2
    \end{bmatrix}
    =
    \begin{bmatrix}
    Q_1 & Q_2
    \end{bmatrix}
    \begin{bmatrix}
    R_{11} & R_{12} \\
    & R_{22}
    \end{bmatrix},
\end{equation}
where we let $\ell = \min(m,n)$ and split $\Pi, Q$, and $R$ into $\Pi_1 \in \R^{m \times k}$, $\Pi_2 \in \R^{m \times (m-k)}$, $Q_1 \in \R^{n \times k}$, $Q_2 \in \R^{n \times (\ell-k)}$, $R_{11} \in \R^{k \times k}$, 
$R_{12} \in \R^{k \times (m-k)}$, and $R_{22} \in \R^{(\ell-k)\times(m-k)}.$ The factorization~\eqref{eq:rankQR} is (strong) rank-revealing if the permutation $\Pi$ ensures that, for all $k$, $R_{11}$ is nearly as well-conditioned as possible and $R_{22}$ is nearly as small as possible. These statements can be formalized~\cite{chandrasekaran1994rank}, though we avoid such a discussion here. In particular, the algorithm of Businger and Golub~\cite{businger1965linear} works well in practice and is available in LAPACK~\cite{lapack,blas3QRCP}---allowing it to be easily incorporated into existing codes. The computational complexity of this method is $\mathcal{O}(nmk)$ when run for $k$ steps.  

Any rank-revealing QR factorization of $A$ immediately yields an ID using $k$ columns with error $\|R_{22}\|_2$ because letting $\I\subset [m]$ be such that $A_{:,\I} = A \Pi_1$ and defining
\begin{equation}
\label{eq:Tmat}
    T = 
    \begin{bmatrix}
    I_k & R_{11}^{-1} R_{12}
    \end{bmatrix}
    \Pi^\top
\end{equation}
we have that 
\[
\|A - A_{:,\I}T\|_2 = \|A - Q_1R_{11}\begin{bmatrix}
    I_k & R_{11}^{-1} R_{12}
    \end{bmatrix}\Pi^T\|_2 = \| A - Q_1 
    \begin{bmatrix}
    R_{11} & R_{12}
    \end{bmatrix}
    \Pi^\top \|_2
    =
    \| R_{22} \|_2.
\]
Choosing $k$ such that $\|R_{22}\|_2\leq \epsilon \|A\|_2$ yields the desired relative error, though in practice it is common to use the diagonal of $R$ as a more efficient proxy for choosing $k.$ Of note, the properties of a rank-revealing QR factorization ensure that the ID is near optimal for a given $k$ and the entries of $T$ are not too large.


\paragraph{Randomized algorithms for computational efficiency}
Ultimately, to use IDs within our compression framework we need to be able to compute them for large matrices --- in particular, ones with significantly more rows than columns. In this setting, prior work~\cite{liberty2007randomized,halko2011finding,armstrong2023structure,dong2023simpler} has shown that that an ID can be effectively computed via a dimension reduction scheme. Canonically, this is accomplished by computing an ID of $GA$ for a suitable random matrix $G$ with many fewer rows than columns (e.g., one with i.i.d. $\mathcal{N}(0,1)$ entries). 

\label{ID}
\section{Compressing Transformers}
\label{Application}
We prune both heads and fully connected layers, since layers of both types make up a significant portion of the FLOPs in a network.  More details on this distribution are provided in the Appendix.  

\subsection{Pruning Fully Connected Layers}

Because fully connected layers form a key part of most transformer architectures, we briefly overview the method from~\citet{chee2022model} that uses IDs to compress them. We consider a simple two layer (one hidden layer) fully connected network written as
\[
f(x;W,U) = U^\top\gamma(W^\top x)
\]
with weights $W\in\R^{d\times n}$ and $U\in\R^{n\times c},$ and activation function $\gamma$. The bias is omitted but can be incorporated by augmenting the data. In this setting, the model is pruned using a collection of i.i.d. unlabeled data $\{x_i\}_{i=1}^m$ collected as the columns of a matrix $X\in\R^{d\times m}.$ Specifically, given an accuracy or size target we compute an ID of $\gamma(W^TX)^\top$ yielding the approximation
\[
\gamma(W^TX) \approx T^\top [\gamma(W^TX)]_{\I,:}
\]
for appropriate $T$ and $\I.$ Because subselection and element-wise activation functions commute, this ID yields the compressed network
\begin{equation}
f(x;W,U) \approx f(x;W,_{:,\I},TU) = U^\top T^\top \gamma((W_{:,\I})^\top x).
\label{eqn:FC}
\end{equation}

This method is able to preserve the network's structure, because in deeper networks $T$ may be folded into the next layer and does not need to be maintained explicitly as a linear layer. Similarly, $W_{:,\I}$ retains the structure of $W$ but has fewer columns (i.e., neurons). The approximation quality in~\eqref{eqn:FC} for all $x$ based on an ID using only finitely many samples depends on both the number of samples and the accuracy tolerance of the ID. \citet{chee2022model} provide theoretical assurances for the method and illustrate that it works quite well with relatively few data points for models with a moderate number of layers.  

However, we find that for deeper networks such as Llama, it is helpful to use the update matrix to additionally correct for errors caused in the previous layers of the network. This can be accomplished by re-running the data through the network, and using a simple least-squares solver to find the optimal correction (which also takes into account errors caused in the input caused by prior layers).  This can be done efficiently using a GPU, but introduces a trade off between the error and the resources used.

\subsection{Pruning Whole Heads}
\label{sec:heads}

We also use an ID to prune entire attention heads.  The structure of the attention heads requires a two-step approach, since we want to constrain our problem to pruning entire blocks of neurons corresponding to an attention head. First, we use an ID to select which heads to keep. The matrix we compute an ID of has columns that are an appropriate ``vectorizatoin'' of each attention head. Once we determine which heads to remove we then solve a least squares problem to compute the correction/interpolation matrix. This second step is crucial to reduce the overall error (as we demonstrate in Section~\ref{Ablation}) because the na\"ive interpolation matrix produced by the first QR factorization is block diagonal---a structure clearly less expressive than the general interpolation matrix we get from solving a least squares problem. Details for these two steps follow.

\paragraph{Step 1: Choosing Heads}
The first step in our process is to choose which heads to keep for each layer. 
Just before the end of an attention block, the outputs from each individual head are concatenated into a single tensor. Typically, this output is then passed through a simple MLP and the result is added to the residual. 
We take the concatenated output of the attention heads, just before applying the MLP, and call this tensor $H$.  Then, we flatten it into an extremely tall and skinny matrix, such that each column in the matrix is all of the outputs that came from a single attention head. By performing a column-pivoted QR on this very tall and skinny matrix, we select which heads to remove---corresponding to the columns that are not selected by the column pivoted QR. Importantly, the permutation matrix computed by a column pivoted QR factorization is invariant to permutations from the left. So, the ``vectorization'' of each attention head can be done somewhat arbitrarily so long as it is done consistently (particularly for step 2).

We begin with a batch of inputs to the attention block, $X = [x_s] \in \R^{m  \times n \times b }$ with $x_s\in\R^{n\times b}$ for $s=1,\ldots,m$, where $m$ is the number of data samples, $b$ is the sample sequence length, and $n$ is the number of neurons in the hidden layers.  Three fully connected layers are applied to this input---the key, query and value layers---with weight matrices $K$, $Q$, and $V\in \R^{n \times n}$.  The tensors $Kx_s$, $Qx_s$, and $Vx_s$ are blocked into the $h$ attention heads, and multiplied together, such that the output of an attention head $H_i \in \R^{m \times  b \times (n/h)}$ where each ``slice'' in the first dimension of $(H_i)_s$ is the output of head $i$ for a given data point $x_s$ computed as $\psi(\phi((K_ix_s)^\top Q_ix_s)(V_ix_s)^\top)$, where $\psi$ and $\phi$ are activation functions.  We stack these along the data dimension to get the output for $X$, $H_i \in \R^{m \times b \times n/h} $.  To calculate the output of an attention block, we typically would concatenate these attention head outputs to get a tensor $H = [H_1, H_2,\ldots, H_h] \in \R^{m  \times n \times b}$. 

 First we want to choose which of our $h$ attention heads to eliminate.  
 For a particular attention head $H_i$, we can think of each entry as an element that we would like to reconstruct given the output of a subset of the heads.  
 Therefore, we ``flatten'' this tensor into a single vector $Z'_i \in \R^{m*b*n/h}$.
 Each input (coming from a single data example) to the attention block contributes $b*n/h$ elements to this vector.    
 We do this for each attention head and let  $Z'_i$ be the columns of a matrix $Z' \in \R^{m*b*n/h\times h} $.  
 We then perform a column pivoted QR on $Z'$, $Z'\Pi = Q'R'$. 
Let $\pi_A \in \mathbb{Z}^{h}$ be the vector that encodes the permutation $\Pi$. We can then permute the outputs of the attention heads as $H' = [H_{\pi_A[1]},H_{\pi_A[2]},\ldots, H_{\pi_A[h]} ]$.  This re-orders the outputs from the heads so that the most important to keep are first and the least important are last.  
In addition, we permute the blocked weight matrices $K$, $Q$, and $V$ according to the same scheme: $K' = [K_{\pi_A[1]},K_{\pi_A[2]},... K_{\pi_A[h]} ]$,$Q' = [Q_{\pi_A[1]},Q_{\pi_A[2]},... Q_{\pi_A[h]} ]$, and $V' = [V_{\pi_A[1]},V_{\pi_A[2]},... V_{\pi_A[h]} ]$. For a selected number of heads to keep $k$ we can then drop the last $(h-k)$ heads from $K'$, $Q'$ and $V'$ to prune the network (but not before computing a correction in the next step).  

\paragraph{Step 2: The Columns-Level Correction:}
The second step is to calculate a correction to account for the heads we have chosen to eliminate.  Here, we have two choices, depending on the available computational budget and the depth of the network.  The computationally cheapest option is to take our tensor $H$ and permute it so that the outputs from the heads we have chosen to eliminate are at the end. We then perform a simple non-pivoted QR factorization to calculate the interpolation matrix. Alternatively, we can take into account errors that have compounded in previous layers by performing a least-squares solve to find the optimal correction. 

To compute a dense correction we start with the permuted output tensor $H'$. 
We can now think of the first two dimensions (the number of data points and the context length) as providing samples for us to reconstruct, and the third dimension (the hidden units in the layer) as the columns we wish to eliminate. 
Therefore, we flatten the first and third dimensions of the $H'$ tensor, yielding a matrix $Z \in \R^{m*b \times n}$. In this setting, each data point in the pruning set contributes $b$ (the context length) samples to this matrix.  
Keeping $k$ heads corresponds to keeping the first $k*n/h$ columns of $Z$ and eliminating the rest.  
Since heads are ordered such that the ones we intend to eliminate are last, we can calculate a QR factorization without pivoting, $Z=QR$, to determine the optimal way to write the eliminated columns as a linear combination of the ones we keep.  
In particular, we partition the rows and columns of $R$ according to the number of columns we want to keep---$n*k/h$---and calculate $T$ according to equation \ref{eq:Tmat} with $\Pi=I.$
We then block $T$ by attention heads getting $T=[T_1, T_2,..., T_k]$, where $T_i \in \R^{n \times n/h}$.  
If $\tilde{\pi}_A$ encodes the inverse permutation from our heads-level QR factorization $\Pi^\top$, we then permute our matrix $T$ according to this block permutation, 
$T_{corrected}=[T_{\tilde{\pi}_A[1]} , T_{\tilde{\pi}_A[2]} ,...,  T_{\tilde{\pi}_A[k]}]
$.  
This gives us the optimal correction to preserve the output of the layer on the pruning data, given our choice of heads to eliminate.  Notably, $T_{corrected}$ is dense, and can be fused into the next layer just like when pruning neurons in the previous section.  Finally, we drop the last $(h-k)$ heads from $K'$, $Q'$ and $V'$ to prune the network.

\subsection{Selecting Pruning Ratios per Layer}
While we have described a compression strategy for fully connected and attention layers given a fixed pruning size or target accuracy, in practice we do not necessarily want to compress all layers equally. 
Some layers may be more sensitive or more important than others and the final size of the compressed network should reflect this---i.e., there are many ways to compress a network to a given overall size and we would like to find a good one.\footnote{Finding the optimal one is likely challenging given the nature of the problem.}  
In particular, we would like to minimize the total error we induce by choosing per-layer compression ratios, and allocate resources based on our given budget.  
We select the allocation choice which minimizes the sum of the error induced by pruning head blocks and fully connected blocks, weighted by the number of FLOPs done by each block type. 

\citet{chee2022model} used an iterative pruning scheme to compress deep convolutional networks.
However, in our setting, we use a simple one-shot technique to determine the per layer sizes for a target FLOPS reduction. 
This is largely for computational efficiency.  It is quite possible that an iterative method could perform better (as found by \cite{ouderaa2024the}), but performing many shots would be costly.    
We choose the per layer sizes to minimize the sum of (estimated) per-layer errors, within the constraint that the sum of flops between the attention heads and neurons is within our budget.
Doing so is reasonably cheap, since we can get all of this information from our pre-computed $R$ matrices for each layer without the need for additional QR factorizations.

To calculate the expected error induced by pruning a given layer, we begin with the error of the ID on the weighted Z, $\|R_{22}\|$. 
The networks we compress contain residual blocks (meaning that the output of a layer is added to the output of the network), which were not addressed by~\citet{chee2022model}.  Importantly, these residual connections change how error propagates through the network.  
In addition, the networks we consider use layernorms~\cite{ba2016layer} instead of batchnorms~\cite{pmlr-v37-ioffe15} and various architectures make different choices about their placement. In fact, we find that the exact placement can impact the magnitude of the intermediary state tensors at different layers in the network.
Based on empirical observation, we use the absolute error ($\|R_{22}\|$) for BERT models, and relative error ($\|R_{22}\|/\|R\|$) for Llama.  
Because inducing errors on the output of the early layers in the network has a larger impact than later layers, we introduce a weighting function which weights the errors at the beginning of the network more. 
This also serves to regularize the allocation process and ensure that the algorithm does overly aggressively prune the early layers.  
We defer details on this weighting function for each network type to the Appendix, and believe that standardizing this process is a good avenue for future work.   

\subsection{Optimizing for Large Networks}
Large decoder models like Llama use much longer sequence lengths (such as 4096 v.s. approx. 100) than the BERT base model and also have much larger weight matrices. 
Typical GPUs may not have enough memory to compute the intermediary state of a sufficiently large pruning set to run the above procedure all at once. 
Moreover, the intermediary state matrix $Z$ may have sufficiently many rows to make the column-pivoted QR factorization itself prohibitively expensive. 
Fortunately, we can use a slight variation on the randomized scheme discussed in Section~\ref{ID} and a simplified version of communication-avoiding rank revealing QR by~\cite{demmel2015communication} to accelerate our method.

\paragraph{Grouping Neurons for Scalability}
Fully-connected layers in models such as BERT have 3027 neurons.  In Llama-2 7B, there are 11008. Like many matrix factorizations, the column pivoted QR factorization scales quadratically in the number of neurons. However, in practice we typically require more data to effectively capture the effects of large layers. This can lead the scaling to become cubic, an issue that we share with many other pruning methods based on matrix factorizations. In addition, column-pivoted QR factorizations cannot be parallelized as easily as non-pivoted QR factorizations or least squares solvers.  

Therefore, we solve this issue by blocking the neurons into groups and determining which neurons per group to keep using a small column-pivoted QR factorization before combining the selected neurons from all groups. This is akin to the strategy in~\cite{demmel2015communication}, albeit without running the full tournament pivoting scheme.

\paragraph{Sketching outputs for Scalability}
To enable our method to scale to large models of practical interest, we use a variant of CountSketch~\cite{clarkson2017low,malik2020fast}. In particular, to compute an ID of $Z$ we first compute an ID of $SZ$ where $S$ is a sparse matrix containing a small number of $\pm 1$ entries. Notably, an ID of $SZ$ provides good information about which columns to keep. We then solve a least squares problem with those columns to computed the ``correction'' matrix. This strategy ensures that the relevant computations can fit on our GPUs.

\label{pruning}
\section{Experimental Results}
\label{Experiment}
We demonstrate the effectiveness of STAT using several different datasets and models, including the BERT and DistilBERT models on the GLUE and SQuAD datasets, and the Llama 7B models on the WikiText2 dataset. The flops ratio is reported as the ratio of flops remaining after compression for a single inference of the encoder or decoder part of the network:  $FLOPS\:Ratio = \frac{New\:Model\:FLOPS\:per\:Inference}{Original\:FLOPS\:per\:Inference}$.  This follows the convention of \cite{rtfp}.  We provide some implementation and hyperparameter details for the slightly different architectures in the Appendix.  All experiments were performed on either a M2 MacBook Pro, or a single A6000 GPU.  

\subsection{BERT and DistilBERT}
\begin{figure}[h!]

\centering
\includegraphics[width=.99\textwidth]{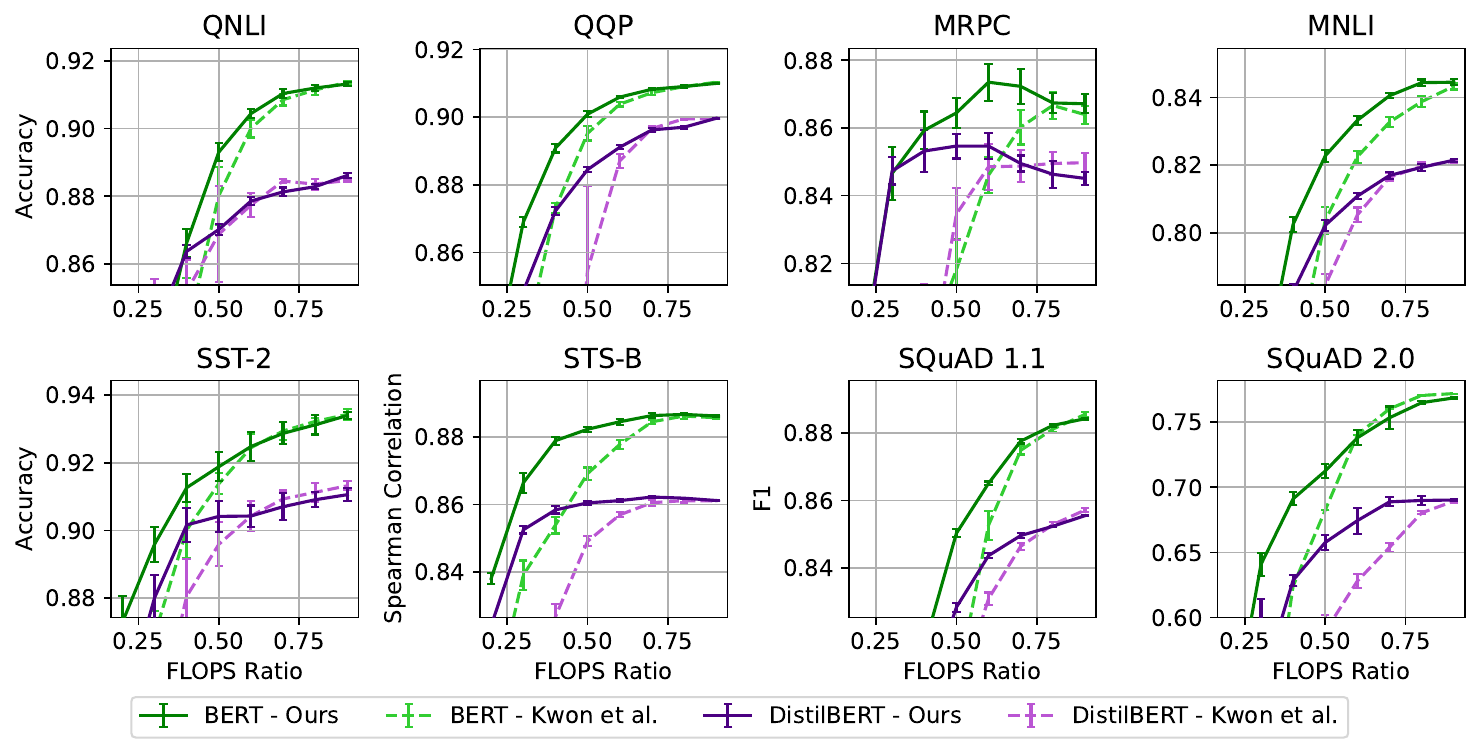}
\caption{Pruning results for BERT and DistilBERT models.  We begin with the exact same base models as \cite{rtfp}.  Error bars are reported as standard deviations across 10 trials of random pruning set selections.  We see that for modest pruning levels, our method performs within error bounds of the baseline method, and at higher compression ratios we substantially outperform \citet{rtfp}.  
}
\label{Bertresult}
\end{figure}

\begin{figure}[h!]
\centering
\includegraphics[width=.99\textwidth]{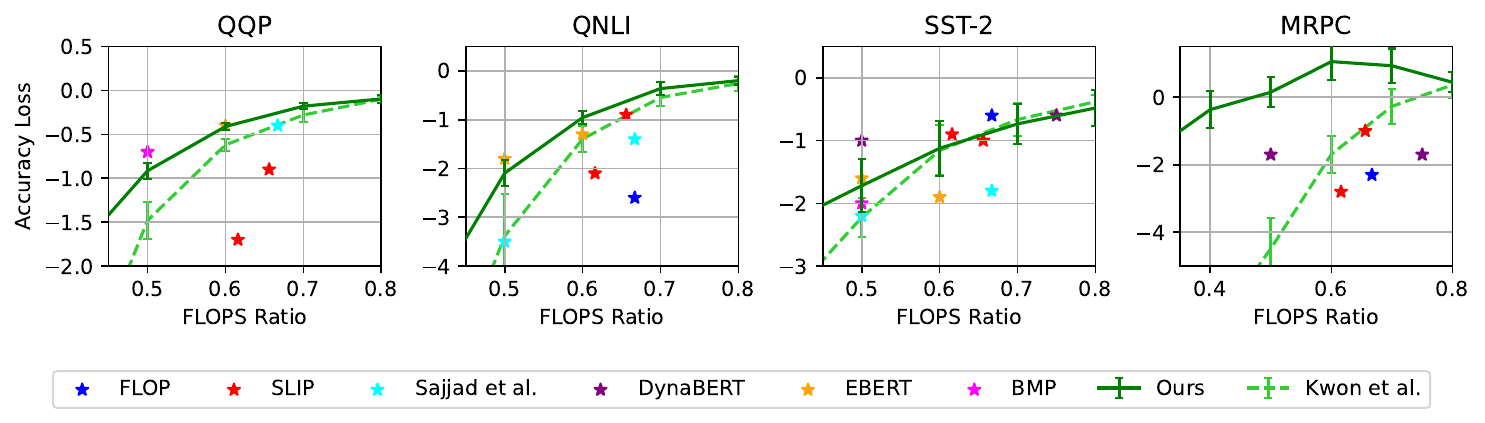}
\caption{Pruning results for BERT model, compared against methods \cite{Wang_2020, lin2020pruning, Sajjad_2023, hou2020dynabert, liu-etal-2021-ebert, lagunas2021block} which include substantial (at least 4 epochs and 5 hours of) fine tuning (stars) and \citet{rtfp}, which does not include fine tuning.  
}
\label{finetuneresult}
\end{figure}

We begin with the pre-trained models provided by \cite{rtfp}. For each desired FLOPS ratio, we prune the network using 512 data examples randomly selected from the training set and error bars represent the standard deviation across 10 different random seeds. Since the tensor size is set by the longest sequence in the batch, we use the attention mask so that we are only reconstructing outputs within the length of each data example.  Figure \ref{Bertresult} gives our results for network sizes ranging from $20\%$ to $90\%$ of the FLOPS of the full size network.  At modest pruning amounts, STAT performs similarly to \cite{rtfp}\footnote{For very small pruning amounts we expect all methods to ``converge'' to the full model performance.} However, as we further reduce the network size STAT significantly outperforms other re-training free methods.  Moreover, in Figure \ref{finetuneresult}, we compare with methods that include substantial (at least 4 epochs, or 5 hours) of fine-tuning on 4 datasets. Interestingly, STAT is often competitive with many of these methods despite not using any fine tuning.

\subsection{Llama}
We also demonstrate the effectiveness of STAT on the decoder-only architecture Llama-2\cite{touvron2023llama}.  We evaluate our pruning results on the smallest model, Llama-2 7B.

For our pruning dataset, we use 256 examples from the C4 training set, each of which consists of sequences of 4096 tokens. This is the same calibration set used by \cite{sun2024a}.  We find this dataset to be more expressive than Wikitext-2.  Results can be found in table \ref{llamaResults}. In order to make the comparisons as fair as possible, we use the perplexity evaluation code and settings from LLM Surgeon \cite{ouderaa2024the} for all methods, even when different evaluation code was used in the original papers.

\begin{table}[h!]
    \centering
    \begin{tabular}{ccccc}
 Method & Size &  WikiText2 & Compression Time& Resources \\
 Baseline &  6.74B$\phantom{^*}$ & 5.12 & &  \\
LLM Surgeon & 5.39B$^*$ & 6.18&17h8m &4xH100 80GB\\
SliceGPT&6.11B$\phantom{^*}$& 6.25&44m&1xH100 80GB\\
SliceGPT&5.70B$\phantom{^*}$& 7.24&44m&1xH100 80GB\\
STAT& 5.87B$\phantom{^*}$&6.43&2hr28m&1xA6000 48GB \\
STAT& 6.30B$\phantom{^*}$&5.62&2hr29m&1xA6000 48GB \\

\end{tabular}
\caption{Results of pruning Llama 2 7B. The pruning ratios given in SliceGPT \cite{ashkboos2024slicegpt} do not necessarily match the number of parameters in the final model---this is because SliceGPT inserts an additional matrix to account for the residual connection in each layer, and thus incurs overhead.  Additionally, when we run on our A6000 setup, SliceGPT takes 3hr6m. (*) LLM Surgeon does not provide the number of model parameters explicitly in their paper the number we provide is computed by multiplying the original model size by their reported ``target size.''}
\label{llamaResults}   
\end{table}

We see that LLM Surgeon produces the best perplexity results for a given size, but at such a high computational cost that it may be difficult to scale to larger models, such as Llama-2 70B.  
This is because the method involves computing gradients for the network in an iterative many shot approach, whereas SliceGPT and STAT are both one-shot gradient-free approaches. 
In the regime we tested, STAT appears better than or equal to SliceGPT, especially since the latter incurs an overhead cost such that the minimum sparsity to keep the model the same size increases the perplexity of the model overall.

\section{Ablation Analysis}
\label{Ablation}
\begin{figure}[h!]
\centering
\begin{subfigure}{.49\textwidth}
  \includegraphics[width=1\linewidth]{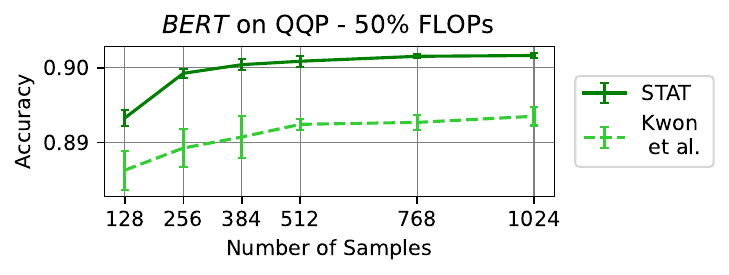} 
  \caption{}
  \label{fig:dataUsage}
\end{subfigure}
\;
\begin{subfigure}{.49\textwidth}
  \includegraphics[width=1\linewidth]{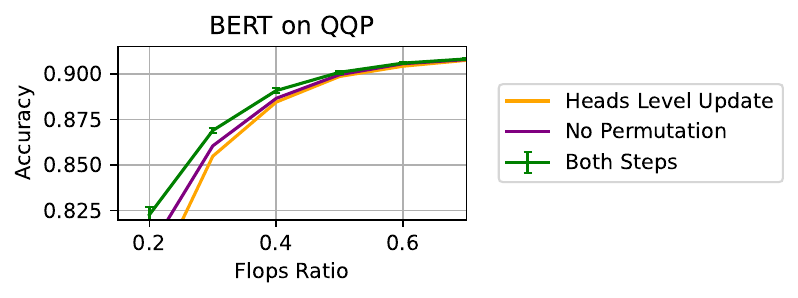} 
  \caption{}
  \label{fig:metrics}
\end{subfigure}
\caption{Left: Accuracy of pruned model on QQP dataset given different amounts of pruning data - We see that accuracy improves until 512 data examples, and then remains fairly steady.  Right: Ablation of the two-step head compression process.  
}
\label{twostep}
\end{figure}
  
\paragraph{Data usage}
We demonstrate the data efficiency of our method compared to \cite{rtfp} by pruning a BERT network that was fine-tuned on the QQP dataset to 50\% FLOPS in Figure \ref{fig:dataUsage}. 
STAT sees improvements in accuracy until roughly 512 data examples are used for the pruning set and exhibits marginal gains when adding more data beyond that point. 
\citet{rtfp} typically uses 2048 data examples for all of their pruning results, and we use 2048 examples for their method in Figure \ref{Bertresult}.

\paragraph{Two step heads}
We use a two-step process to compress attention blocks.  First, we select heads to prune by doing a column-pivoted QR on the heads level, and second we re-calculate an update by performing an non-pivoted QR decomposition at the columns level. Figure \ref{fig:metrics} demonstrates that the method requires both steps to reach maintain full accuracy.

\section{Limitations and Future Work}
STAT is a method to compress transformer networks without fine tuning that achieves state of the art results on several combinations of models and datasets. 
However, there remains several limitations to the method and avenues for future work.  
First, we believe that we have not exhausted the ability to improve the performance of our method by increasing the pruning set size for the Llama models, though this would increase the time to compress the networks. 
Perhaps the greatest current limitation of our method is the the need to slightly adapt the per-layer error metric computation for different network architectures.  
We hypothesize that the placement of the layer-norm in a network changes the sensitivity of the network to changes in outputs from different layers, and a systematic study is needed to confirm this hypothesis and determine a path forward. 
Some preliminary exploration of these points can be found in the Appendix and we plan to pursue these avenues in future work.  

\section*{Acknowledgements}
M.F. and A.D. were supported by the SciAI Center, funded by the Office of Naval Research under Grant Number N00014-23-1-2729. C.D. was supported by Google and NSF CAREER-2046760.

{
\small
\bibliographystyle{plainnat}
\bibliography{Transformer.bib}
}


\appendix

\section{Appendix / supplemental material}
\subsection{Wall Clock Time Analysis on Consumer Hardware}
We measure the wall clock time to evaluate the performance of a BERT network on the QQP test dataset at various levels of compression.  All networks are run on the MPS cores of a 2023 MacBook M2 Pro chip with 32 GB of memory.  Our measurement of wall clock time begins after the model is loaded into memory, but includes loading data into memory and the embedding.  We find that the entire evaluation pipeline is 25\% faster when we reduce the encoder size by 50\%.   

\begin{table}[h!]
    \centering
    \begin{tabular}{cc}
        Model Size (Flops ratio) & Time to evaluate test set (s) \\
        1.0 & 162 $\pm$ 3\\
        0.5 & 120 $\pm$ 2\\
        0.25 & 96 $\pm$ 2
    \end{tabular}
\end{table}

\subsection{Low-Rank Decomposition of Attention Head Matrices}
\begin{figure}[h!]

\centering
\includegraphics[width=.6\textwidth]{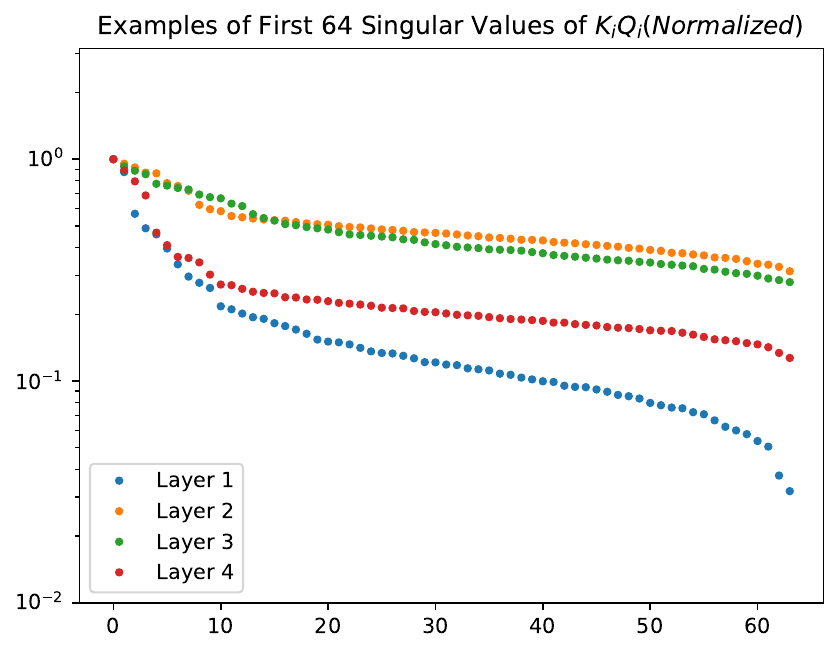}
\caption{First 64 normalized singular values of the $K_iQ_i$ matrix for 4 heads randomly selected from the first 4 layers of a network.  
}
\label{singval}
\end{figure}

In addition to pruning entire attention heads, we attempted to compress the matrices within attention heads using the SVD. Specifically, we replaced the matrices for each the key and query of each head, $K_i$ and $Q_i$, with the singular value decomposition. Using the SVD $K_iQ_i=U_i\Sigma_i V_i^\top$ we can replace the weights with $K_i \rightarrow U_i^{(r)} \Sigma_i^{(r)}$, and $Q_i \rightarrow (V_i^{(r)})^\top,$ where $(\cdot)^{(r)}$ denotes either truncating a matrix to its first $r$ columns (as for $U_i$ and $V_i)$ or upper left $r\times r$ block (as for $\Sigma_i$). The truncation parameter $r$ is chosen to ensure the error introduced by the dimension reduction is amenable to the overall accuracy goal. However, we found that compressing the attention heads in this manner was not particularly
useful for fine-tuning-free compression because the singular values of $K_iQ_i$ tend to decay quite slowly---as illustrated in Figure~\ref{singval}.  For example, compressing the attention heads from 64 to 55 neurons on a BERT model trained on QQP decreased the accuracy from 91.0\% to 90.6\%, despite the fact that doing so only decreased the FLOPS in the network by a few percent. While we may be able to prune some heads more than others, creating mixed-size heads in layers introduces more significant engineering complications than creating layers of different sizes or numbers of heads.

\subsection{$\|R_{22}\|$ Tracks the Layer Error}

\begin{figure}[h!]

\centering
\includegraphics[width=.65\textwidth]{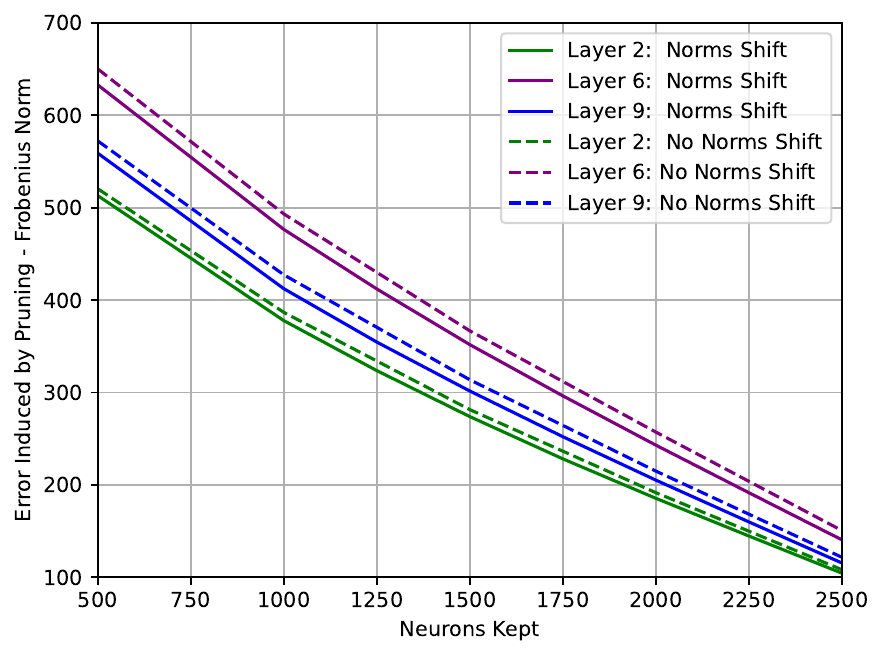}
\caption{We prune 3 layers in the network and show the difference in the error induced on the output of the layer on the pruning set.  Shifting the norms from the next layer improves the error slightly. 
}
\label{Norms}
\end{figure}

In contrast to~\citet{chee2022model}, we use the norm of the next layer to re-weight the rows in the matrix $Z$ before computing the interpolative decomposition.  This slightly improves the preservation of the per layer error, as shown in Figure \ref{Norms}.

\begin{figure}[h!]

\centering
\includegraphics[width=.65\textwidth]{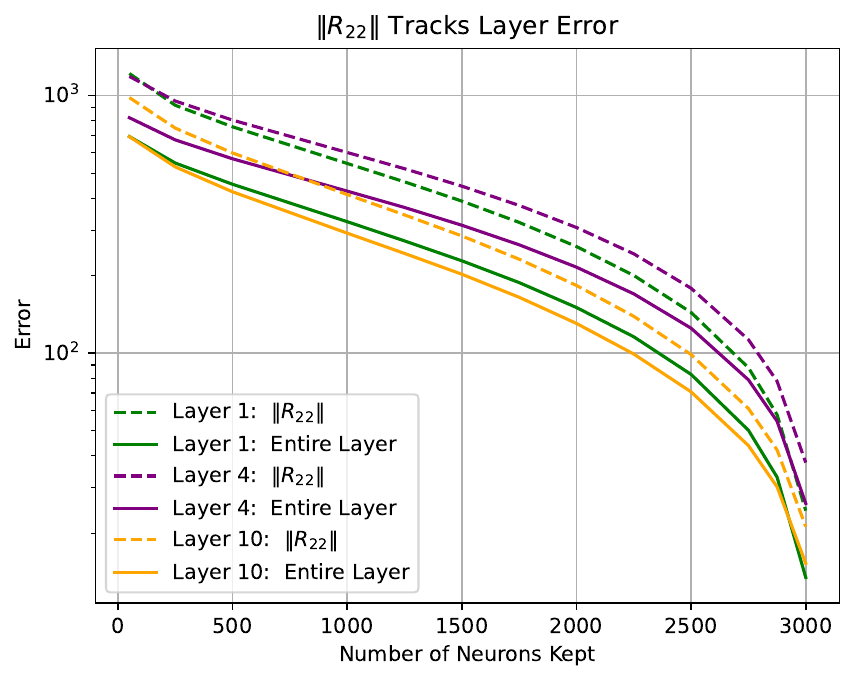}
\caption{We prune 3 layers in the network (one in the beginning, middle and end) to different numbers of neurons, and measure both our error metric $\|R_{22}\|$ and the actual error in the output of the layer on the pruning dataset.  
}
\label{R22}
\end{figure}

More importantly, Figure~\ref{R22} shows that we can use $\|R_{22}\|$ as a proxy for the error introduced when we prune a layer. To illustrate this we prune some of the fully connected blocks to different numbers of neurons and measure both $\|R_{22}\|_F$ and the Frobenius norm of the error introduced in the output of the entire layer on the pruning data.  While $\|R_{22}\|$ does not take into account any activation function of the last fully connected layer in the fully connected block, we see that the two error metrics track each other fairly closely. Moreover, $\|R_{22}\|_F$ is always an overestimate of the error in this case, meaning we are unlikely to introduce more error than desired (instead we may not quite prune as much as possible).

\subsection{Layer norms and regularization of network size}

\begin{figure}[h!]
\centering
\begin{subfigure}{.47\textwidth}
  \includegraphics[width=1\linewidth]{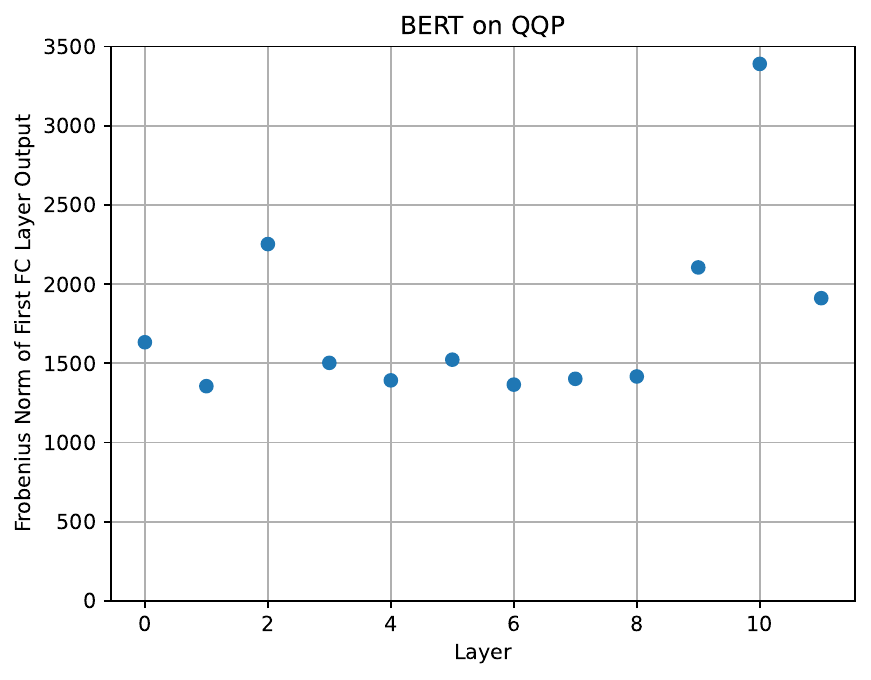} 
  \caption{}
  \label{fig:BERTNorms}
\end{subfigure}
\;
\begin{subfigure}{.49\textwidth}
  \includegraphics[width=1\linewidth]{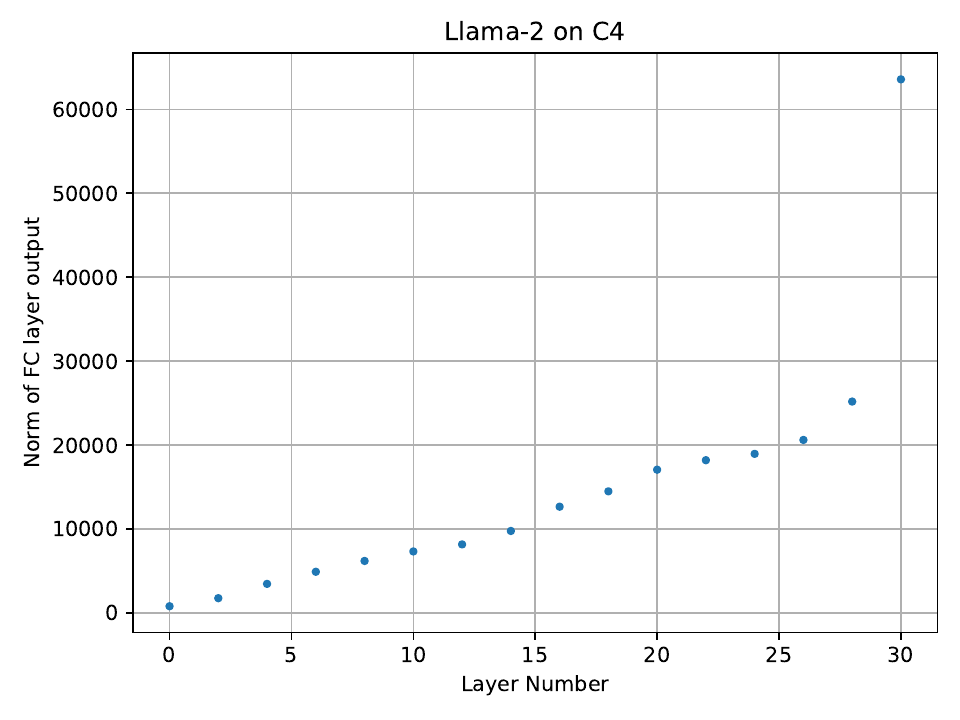} 
  \caption{}
  \label{fig:LlamaNorms}
\end{subfigure}
\caption{ Frobenius norms of the output of the first fully connected layer in a fully connected block as it passes through the network for the BERT base model and Llama-2 7B.  For the BERT model, there is not a clear relationship between where the layer is in the network and the Frobenius norm of its output, whereas for Llama-2 the norm grows approximately linearly.
}
\label{norms}
\end{figure}

BERT and Llama make different choices about where to place the layer norm within the network. BERT applies a layernorm to the hidden state of the network after adding the output of the layer to the residual connection.  Llama applies the layernorm to the input of the attention heads and fully connected layers.Figure \ref{norms} shows that the norm of the output of the first fully connected layer in a fully connected block of a BERT model does not appear to be strongly correlated with the layer's position in the network.  In contrast, the norm of the output appears to be roughly quadratically related to the layer number for a Llama model.  

Because of these architectural discrepancies we make slightly different choices in how to apply our method to BERT and OPT models. In a BERT model we use the absolute error of the network to determine the per-layer sizes. However, this can cause STAT to over-prune the first few layers in the network, which has a disproportionate effect on the overall error since outputs of the early layers affect the inputs of the later layers. 
Therefore, we apply a weighting function to the expected error of the layers when determining how much to prune each layer: if $l$ is the layer's position in the network ($l=1$ for the first layer), then the weighting is given by:  $\sqrt{l+1}+1$.  
This was chosen empirically based on a single seed for BERT on the QQP dataset, and applied to all combinations of BERT, DistilBERT, and datasets.

For Llama, we notice that the norm of the outputs increases linearly, and measured the norm of the gradients for early layers and notice that they increase linearly relative to the size of the norm of the outputs of the layer for general networks with the llama architecture.  Therefore, we apply a weighting function of l+50 to the expected relative error.


\subsection{The computational cost of pruning}
The computational cost of our algorithm is dominated by the cost of performing $3*N$ QR decompositions, where $N$ is the number of layers in the network. 
However, the overall cost is generally still relatively modest, and much less than fine-tuning.  
For example, pruning an entire BERT model on a Mac Book using an M2 chip takes 4 minutes and 25 seconds on the qqp dataset using 512 data examples for the pruning set. 
Pruning a larger OPT model (for example, 1.3B) takes 2.643 hours (About 2 hours and 38 minutes) on average using a single RTX 3090 GPU with 24 GB of RAM.
In this setting the run-time is dominated by the cost of saving intermediate steps to disk. 
Most of the pruning process can be easily parallelized, since the contribution to each layer's output matrices can be computed for each data point independently and once the intermediary states for each data point are collected the QR factorizations for each layer can be done independently. Additionally, our method can produce models of many different sizes while only computing these QR factorizations once, since the $R$ matrices are the same for all FLOPs ratios and all that changes is the chosen truncation parameter $k$.   

\subsection{Experimental Design}
Design choices for our method were made using BERT and a single random seed to prune it on the qqp dataset. We then validated using the other 7 datasets, DistilBERT model, and different random seeds.  We made all optimizations for Llama 2-7B using random seed 0 for data selection, and validated using different slices of the pruning dataset.

\subsection{Model Correlation }
\begin{figure}[h!]

\centering
\includegraphics[width=.99\textwidth]{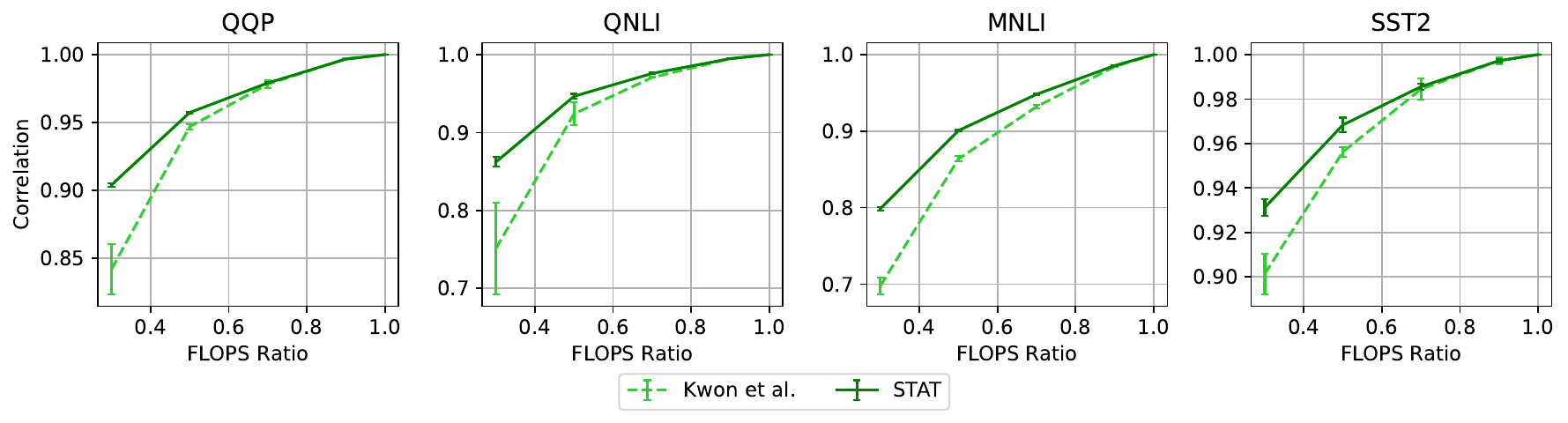}
\caption{Model correlation results for the full-sized BERT model on 4 datasets.   
}
\label{Corrresult}
\end{figure}

Chee et al. \cite{chee2022model} noted that pruned models may not preserve desirable properties of the full size model if excessive fine tuning is done. They propose ``model correlation", or the percentage of test samples the the two models agree on, as a rough measure of how well the pruning method preserves the original model.  We believe that this metric is less useful here---we focus on compressing the network and maintaining accuracy without fine tuning, so it is unlikely that we would succeed in doing so without preserving the ``model correlation.''  However, in Figure~\ref{Corrresult} we provide results on this metric for 4 data sets on the BERT model.  We see that at all levels of compression, STAT preserves the correlation between the original and pruned model better than \citet{rtfp}.

\end{document}